\documentclass[conference]{IEEEtran}
\IEEEoverridecommandlockouts
\usepackage{cite}
\usepackage{amsmath,amssymb,amsfonts}
\usepackage{algorithmic}
\usepackage{graphicx}
\usepackage{multirow}
\usepackage{subfigure}
\usepackage{textcomp}
\usepackage{xcolor}
\def\BibTeX{{\rm B\kern-.05em{\sc i\kern-.025em b}\kern-.08em
    T\kern-.1667em\lower.7ex\hbox{E}\kern-.125emX}}
\begin{document}

\title{PVContext: Hybrid Context Model for Point Cloud
Compression
% {\footnotesize \textsuperscript{*}Note: Sub-titles are not captured for https://ieeexplore.ieee.org  and
}

\author{\IEEEauthorblockN{1\textsuperscript{st} Guoqing Zhang}
\IEEEauthorblockA{\textit{dept. Computer Science and Technology} \\
\textit{Harbin Institute of Technology}\\
Harbin, China \\
hitcszgq@stu.hit.edu.cn}
\and
\IEEEauthorblockN{2\textsuperscript{nd} Wenbo Zhao}
\IEEEauthorblockA{\textit{dept. Computer Science and Technology} \\ 
\textit{Harbin Institute of Technology}\\
Harbin, China \\
wbzhao@hit.edu.cn}
\and
\IEEEauthorblockN{3\textsuperscript{rd} Jian Liu}
\IEEEauthorblockA{\textit{dept. Computer Science and Technology} \\ 
\textit{Harbin Institute of Technology}\\
Harbin, China \\
hitcslj@stu.hit.edu.cn	}
\and
\IEEEauthorblockN{4\textsuperscript{th} Yuanchao Bai}
\IEEEauthorblockA{\textit{dept. Computer Science and Technology} \\ 
\textit{Harbin Institute of Technology}\\
Harbin, China \\
yuanchao.bai@hit.edu.cn}
\and
\IEEEauthorblockN{5\textsuperscript{th} Junjun Jiang}
\IEEEauthorblockA{\textit{dept. Computer Science and Technology} \\ 
\textit{Harbin Institute of Technology}\\
Harbin, China \\
jiangjunjun@hit.edu.cn	}
\and
\IEEEauthorblockN{6\textsuperscript{th} Xianming Liu}
\IEEEauthorblockA{\textit{dept. Computer Science and Technology} \\ 
\textit{Harbin Institute of Technology}\\
Harbin, China \\
csxm@hit.edu.cn	}
}

\maketitle

\begin{abstract}
Efficient storage of large-scale point cloud data has become increasingly challenging due to advancements in scanning technology. Recent deep learning techniques have revolutionized this field; However, most existing approaches rely on single-modality contexts, such as octree nodes or voxel occupancy, limiting their ability to capture information across large regions. In this paper, we propose PVContext, a hybrid context model for effective octree-based point cloud compression. PVContext comprises two components with distinct modalities: the Voxel Context, which accurately represents local geometric information using voxels, and the Point Context, which efficiently preserves global shape information from point clouds. By integrating these two contexts, we retain detailed information across large areas while controlling the context size. The combined context is then fed into a deep entropy model to accurately predict occupancy. Experimental results demonstrate that, compared to G-PCC, our method reduces the bitrate by 37.95\% on SemanticKITTI LiDAR point clouds and by 48.98\% and 36.36\% on dense object point clouds from MPEG 8i and MVUB, respectively.
\end{abstract}


\section{Introduction}
Point clouds play a vital role in representing 3D objects and scenes. With the rapid advancement of sensor technology, their use has expanded across domains such as autonomous driving, remote sensing, and virtual reality. However, these technological advancements have also dramatically increased both the volume and acquisition rate of point cloud data, posing substantial challenges for transmission and storage. This growing challenge has, in turn, spurred extensive research into point cloud compression.

Due to the irregular structure of point clouds, most existing point cloud compression methods first convert them into regular structures. These methods then estimate occupancy probabilities by constructing context models, achieving compression through entropy encoding. For instance, Draco~\cite{draco} converts point clouds into a KD-Tree before applying compression, while G-PCC~\cite{gpcc} partitions the point cloud into an octree and utilizes the occupancy of compressed nodes as contexts. However, constrained by the limitations of manually designed context models, these methods typically rely on a small number of neighboring nodes, which restricts their overall performance.

Recent deep learning-based compression methods~\cite{tu2019point,biswas2021muscle,nguyen2021multiscale} leverage neural network-driven entropy models to automate information extraction, eliminating the need for handcrafted contexts. Nguyen \textit{et al.}\cite{voxeldnn} employ masked 3D CNNs for voxel-based point cloud compression, while VoxelContext\cite{que2021voxelcontext} enhances performance by utilizing local voxel information from closely related nodes. However, as the context size increases, the volume of the 3D context grows substantially, leading to a significant rise in computational cost. To address this, methods like OctSqueeze~\cite{huang2020octsqueeze} and OctAttention~\cite{fu2022octattention} transform octree nodes into linear representations, extending the context to thousands of nodes. However, this transformation also results in the loss of local structural details, reducing compression efficiency.

To address this challenge, we propose PVContext, a hybrid context model designed for efficient large-scale information representation. PVContext integrates two contexts with distinct modalities: the Voxel Context, which preserves local structures using voxel blocks from precursor nodes, and the Point Context, which maintains global shape information from reconstructed ancestor point clouds. By combining these two contexts, we effectively control the growth of context volume while retaining detailed information over large regions. Additionally, we introduce a hybrid entropy model that extracts and fuses features from each context to predict occupancy probabilities. Experimental results demonstrate that PVContext delivers superior compression performance for both LiDAR and object point clouds. The main contributions of our work are as follows:
\begin{itemize} 
\item We propose a novel hybrid context for octree-based point cloud compression that expands the context range while preventing information in large regions and avoiding rapid volume growth.  
\item We propose a novel network that employs distinct encoders to extract features from each context and perform efficient feature fusion. 
\item The proposed method achieves superior compression performance across LiDAR and object point clouds.
\end{itemize}  

\begin{figure*}[ht]
\centering
\includegraphics[width=\textwidth]{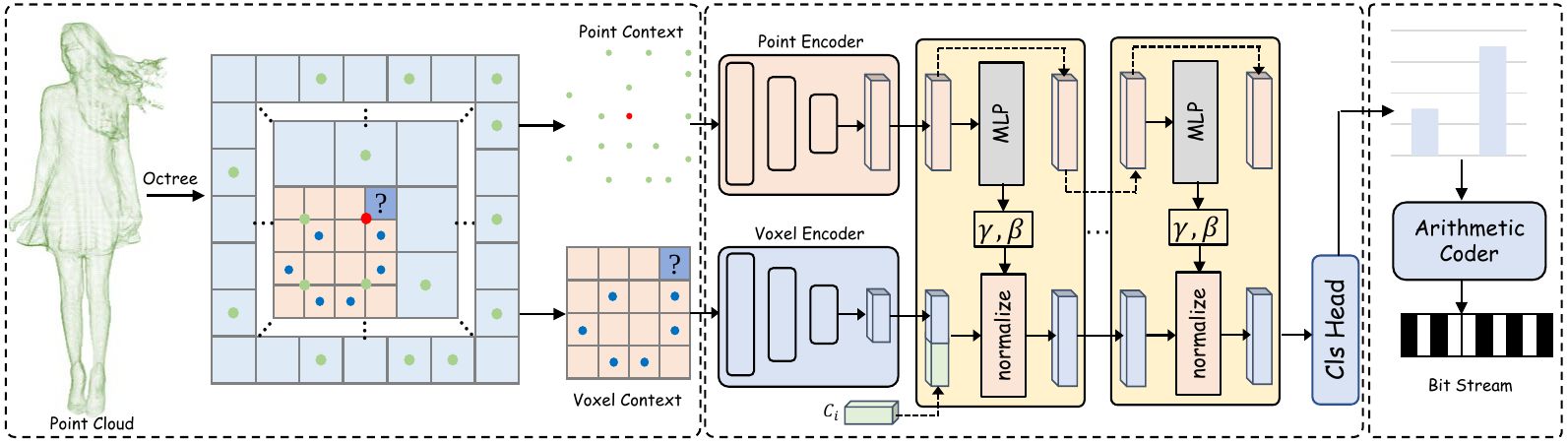} 
\caption{The overview of our method. The input point cloud is first processed using an octree. To predict the occupancy state of the current node (blue), we form its precursor encoded nodes (orange) as \textit{Voxel Context}, and the neighbor points (light blue) of its parent node (red) as \textit{Red Context}. These context are then fed to a encoder-decoder based network, which predicts the occupancy probability of the current node. Finally, arithmetic encoding is used to compress the octree into a compressed bitstream based on the estimated state distribution.} 
\label{Fig:overview}
\end{figure*} 

\section{Method}
\subsection{Overall Framework}
Let $\mathcal{X} = {p_i} \in \mathbb{R}^{n \times 3}$ represent the raw point cloud. Our objective is to represent $\mathcal{X}$ as a sequence of symbols ${s_i}$ within an octree structure and compress this sequence using a context-based entropy coder. To efficiently represent information over a large range, we propose a novel hybrid context model, termed PVContext. As illustrated in Fig.~\ref{Fig:overview}, PVContext integrates two distinct contexts with different modalities: the Voxel Context, which captures local structures through voxel blocks from precursor nodes, and the Point Context, which provides global shape priors from reconstructed point clouds in ancestor layers. This combination enables the capture of large-scale information while controlling the growth of context volume, resulting in more efficient point cloud compression.

\subsection{Octree Structure}
We begin by serializing $\mathcal{X}$ through the construction of an octree. Let the bit precision of $\mathcal{X}$ be denoted as $N$. The 3D bounding box of $\mathcal{X}$ is used as the root node of the octree, which is recursively subdivided into eight child nodes for each non-empty node, until a preset depth $D$ is reached, where $D \leq N$. Once the octree is constructed, it is traversed in a breadth-first order, with the occupancy of each node represented by a symbol $s_i$. The choice of $D$ determines the precision of the representation: when $D < N$, the representation is lossy, while increasing $D$ allows the octree to capture finer details. When $D = N$, $\mathcal{X}$ is represented losslessly.

\subsection{PVContext}
According to information theory \cite{shannon2001mathematical}, given a symbol sequence $\boldsymbol{\mathbf{s}} = [s_1, s_2, \dots, s_n]$, the theoretical lower bound of the bitrate is determined by $\mathop{\mathbb{E}}_{s\sim P}[-\log_{2} P(\boldsymbol{\mathbf{s}})]$, where $P(\mathbf{s})$ represents the true underlying distribution. Since $P(\mathbf{s})$ is unknown during compression, we aim to estimate a distribution $Q(\mathbf{s})$ that minimizes the cross-entropy $\mathop{\mathbb{E}}_{s\sim P}[-\log_{2} Q(\boldsymbol{\mathbf{s}})]$. The closer the estimated distribution $Q(\mathbf{s})$ is to the true distribution $P(\mathbf{s})$, the lower the resulting bitrate.

To accurately estimate $P(\mathbf{s})$, we introduce PVContext, a novel approach that combines features from different octree levels and leverages the strengths of both point cloud and voxel modalities. As illustrated in Fig.~\ref{Fig:overview}, PVContext comprises two main components: the Voxel Context $\boldsymbol{h}^{\text{vox}}$, which captures local structures in precursor encoded nodes, and the Point Context $\boldsymbol{h}^{\text{pc}}$, which extracts global shape information from ancestor layers. Specifically, for the current node $n_i$, the generation process of these two contexts is as follows:

\noindent\textbf{Voxel Context.} The precursor encoded nodes play a crucial role in probability estimation, as they provide information about the local structure near the current node. However, directly using the occupancy of these nodes can result in a loss of structural detail. For the current node $n_i$, we construct a local voxel block $\boldsymbol{h}^{\text{vox}}$ with a size of $4 \times 4 \times 4$ and traverse the precursor nodes to obtain the occupancy status of each voxel. The $\boldsymbol{h}^{\text{vox}}$ is then used as the Voxel Context, which preserves local structural information.

\noindent\textbf{Point Context.} Due to the breadth-first traversal order, only a partial shape prior can be accessed from the precursor encoded nodes. To capture the full shape prior, we extract it from the ancestor layer, which has already been fully compressed. As voxel size increases rapidly with depth, we propose using point clouds to efficiently preserve the shape prior across larger regions. Specifically, for the current node $n_i$ in octree level $d$, the reconstructed point cloud of the ancestor layer is denoted as $\mathcal{X}_{d-1}$, and the proposed Point Context $\boldsymbol{h}_i^{pc}$ is constructed by selecting the $K$ nearest neighbors of $n_i$ from $\mathcal{X}_{d-1}$. 

Finally, the entropy model can be formally described as follows:  
\begin{equation}
Q(\mathbf{s})=\prod_{i} q(s_i\mid \boldsymbol{h}_i^{pc}, \boldsymbol{h}_i^{vox}, \boldsymbol{c}_i, \boldsymbol{\theta})
\end{equation}
where $\boldsymbol{c}_i$ is the coordinate of $n_i$, $\boldsymbol{\theta}$ is the model parameter. 

\section{Hybrid Entropy Model}
As depicted in Fig.~\ref{Fig:overview}, our proposed entropy model adopts an encoder-decoder architecture \cite{sutskever2014sequence}, where $\boldsymbol{h}_i^{pc}$ and $\boldsymbol{h}_i^{vol}$ are fed into separate encoders for feature extraction. The decoder then combines these extracted features to predict the occupancy probability of the node.

\subsection{Network Architecture.}\label{sec:Architecture} 
\noindent\textbf{Encoder.} The encoder consists of a point cloud encoder $f_p$ and a voxel encoder $f_v$. We use PointNet \cite{qi2017a} as $f_p$: $\boldsymbol{h}_i^{pc}$ passes through four fully connected (FC) layers (with dimensions 64, 64, 128, and 256) followed by max-pooling and multiscale feature fusion. All layers utilize batch normalization and ReLU activation, ultimately producing a 1024-dimensional feature vector, $\boldsymbol{E}_i^{pc}$.

\begin{equation}
\boldsymbol{E}_i^{pc}=f_{p}(\boldsymbol{h}_i^{pc})
\end{equation}

Similarly, we construct $f_v$ by cascaded 3D convolutions, which mirrors the structure of $f_v$, and produce a 512-dimensional feature vector $\boldsymbol{E}_i^{vox}$:
\begin{equation}
\boldsymbol{E}_i^{vox}=f_{v}(\boldsymbol{h}_i^{vox})
\end{equation}

\noindent\textbf{Decoder.} The decoder $f_{dec}$ receives the feature vectors $\boldsymbol{E}_i^{pc}$, $\boldsymbol{E}_i^{vox}$, and the coordinates $\boldsymbol{c}i$ of $n_i$ to predict the occupancy probability of $q(s_i)$, which can be writed by:

\begin{equation}
q(s_i) = f_{dec}(\boldsymbol{E}_i^{pc}, \boldsymbol{E}_i^{vox}, \boldsymbol{c}_i)
\end{equation}  

The proposed decoder $f_{dec}$ first employs a coordinate embedding layer to transform $\boldsymbol{c}_i$ into a 512-dimensional vector, $\boldsymbol{E}_i^{coor}$. This embedding is then concatenated with $\boldsymbol{E}_i^{vox}$ and $\boldsymbol{E}_i^{pc}$. The concatenated features are passed through five residual blocks, each equipped with Conditional Batch Normalization \cite{de2017modulating}, followed by a sigmoid-activated fully connected (FC) layer, which outputs $q(s_i)$. The input and output dimensions of the residual blocks remain the same.

\subsection{Loss Function}\label{sec::loss}
We employ the binary cross-entropy loss to measure the difference between the predicted occupancy state and ground truth:
\begin{equation}
L_{BCE}=- \sum_{i} p(s_i)\log q (s_i)
\end{equation}
where $p(s_i)$ is the ground-truth occupancy state of node $s_i$, and $q(s_i)$ is the predicted occupancy state by the model.

\begin{figure*}[!ht]
\centering 
\subfigure {\includegraphics[scale=0.33]{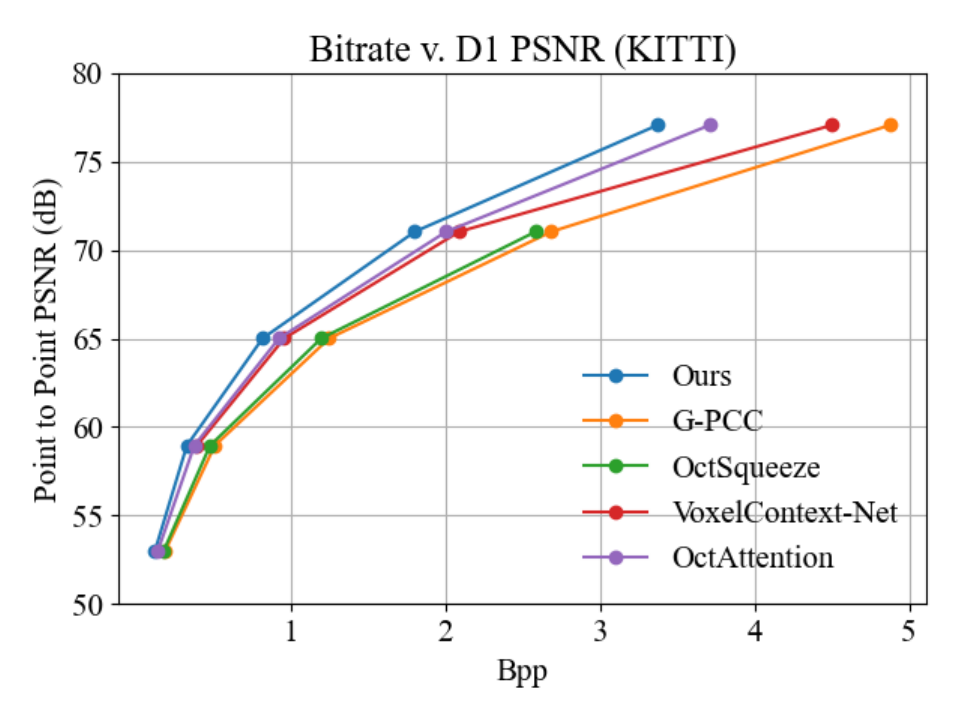}}
\subfigure {\includegraphics[scale=0.33]{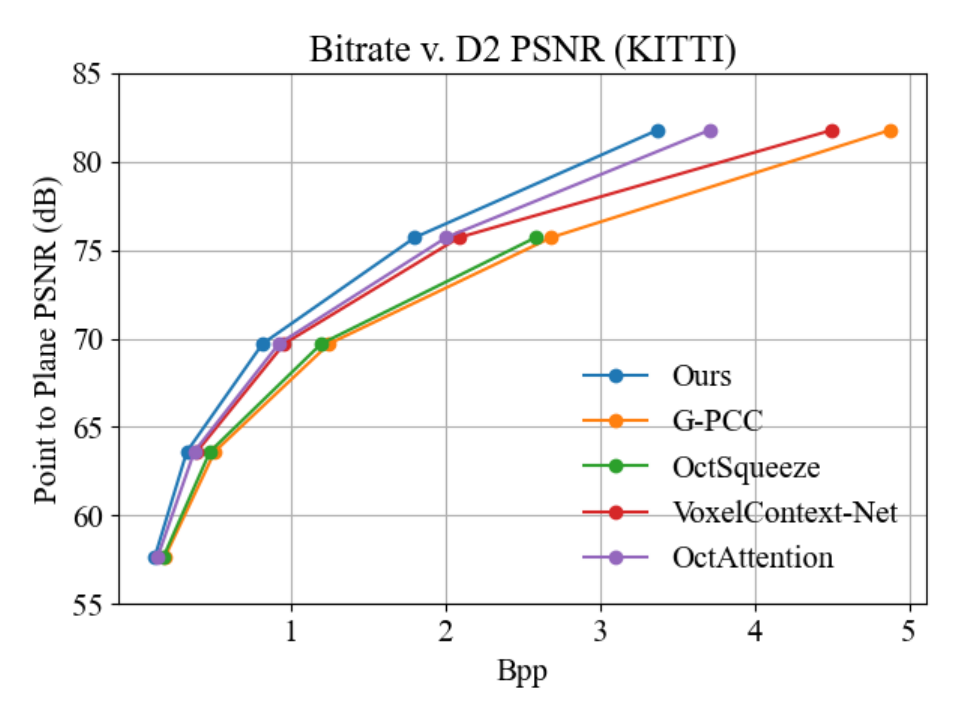}}
\subfigure {\includegraphics[scale=0.33]{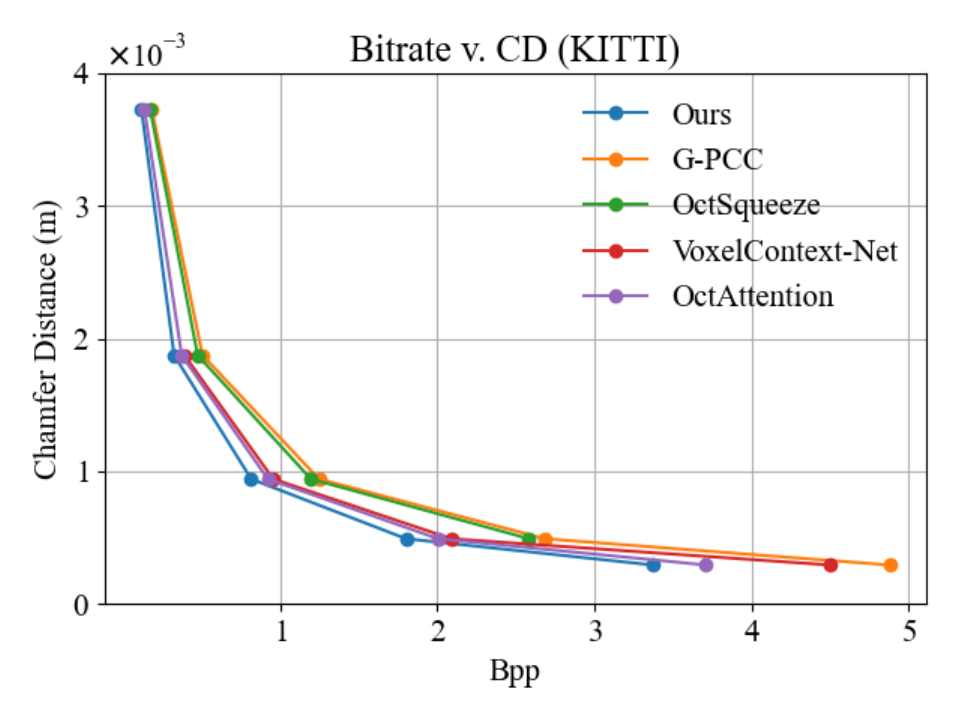}}
\caption{Results of different methods on SemanticKITTI at different bitrates.}
\label{Fig:FIGKITTI}
\end{figure*}

\begin{table*}[ht]
\centering
\caption{Average bits per point (bpp) results of different methods on MVUB and MPEG 8i.}
\resizebox{0.85\linewidth}{!}{
\begin{tabular}{|cccccccc|}
\hline
\multicolumn{1}{|c|}{Point Cloud}    & \multicolumn{1}{c|}{Frames} & \multicolumn{1}{c|}{G-PCC}     & \multicolumn{1}{c|}{VoxelDNN} & \multicolumn{1}{c|}{MSVoxelDNN} & \multicolumn{1}{c|}{OctAttention} & \multicolumn{1}{c|}{Ours}               & Gain over G-PCC \\ \hline
\multicolumn{8}{|c|}{Microsoft Voxelized Upper Bodies (MVUB)}                                                                                                                                                                                                         \\ \hline
\multicolumn{1}{|c|}{Phil9}          & \multicolumn{1}{c|}{245}    & \multicolumn{1}{c|}{1.23}      & \multicolumn{1}{c|}{0.92}     & \multicolumn{1}{c|}{-}          & \multicolumn{1}{c|}{0.83}         & \multicolumn{1}{c|}{\textbf{0.79}}      & -35.77\%        \\ \hline
\multicolumn{1}{|c|}{Phil10}         & \multicolumn{1}{c|}{245}    & \multicolumn{1}{c|}{1.07}      & \multicolumn{1}{c|}{0.83}     & \multicolumn{1}{c|}{1.02}       & \multicolumn{1}{c|}{0.79}         & \multicolumn{1}{c|}{\textbf{0.72}}      & -32.71\%        \\ \hline
\multicolumn{1}{|c|}{Ricardo9}       & \multicolumn{1}{c|}{216}    & \multicolumn{1}{c|}{1.04}      & \multicolumn{1}{c|}{0.72}     & \multicolumn{1}{c|}{-}          & \multicolumn{1}{c|}{0.72}         & \multicolumn{1}{c|}{\textbf{0.68}}      & -34.61\%        \\ \hline
\multicolumn{1}{|c|}{Ricardo10}      & \multicolumn{1}{c|}{216}    & \multicolumn{1}{c|}{1.07}      & \multicolumn{1}{c|}{0.75}     & \multicolumn{1}{c|}{0.95}       & \multicolumn{1}{c|}{0.72}         & \multicolumn{1}{c|}{\textbf{0.65}}      & -39.25\%        \\ \hline
\multicolumn{1}{|c|}{Average}        & \multicolumn{1}{c|}{-}      & \multicolumn{1}{c|}{1.10}      & \multicolumn{1}{c|}{0.81}     & \multicolumn{1}{c|}{0.99}       & \multicolumn{1}{c|}{0.76}         & \multicolumn{1}{c|}{\textbf{0.70}}                   &    -36.36\%              \\ \hline
\multicolumn{8}{|c|}{8i Voxelized Full Bodies (MPEG 8i)}                                                                                                                                                                                                              \\ \hline
\multicolumn{1}{|c|}{Loot10}         & \multicolumn{1}{c|}{300}    & \multicolumn{1}{c|}{0.95}      & \multicolumn{1}{c|}{0.64}     & \multicolumn{1}{c|}{0.73}       & \multicolumn{1}{c|}{0.62}         & \multicolumn{1}{c|}{\textbf{0.48}}      & -49.47\%        \\ \hline
\multicolumn{1}{|c|}{Redandblack10}  & \multicolumn{1}{c|}{300}    & \multicolumn{1}{c|}{1.09}      & \multicolumn{1}{c|}{0.73}     & \multicolumn{1}{c|}{0.87}       & \multicolumn{1}{c|}{0.73}         & \multicolumn{1}{c|}{\textbf{0.59}}      & -45.87\%        \\ \hline
\multicolumn{1}{|c|}{Boxer10}      & \multicolumn{1}{c|}{1}      & \multicolumn{1}{c|}{0.94} & \multicolumn{1}{c|}{-}   & \multicolumn{1}{c|}{0.70}     & \multicolumn{1}{c|}{0.59}    & \multicolumn{1}{c|}{\textbf{0.45}} & -52.12\%        \\ \hline
\multicolumn{1}{|c|}{Thaidancer10} & \multicolumn{1}{c|}{1}      & \multicolumn{1}{c|}{0.99} & \multicolumn{1}{c|}{-}   & \multicolumn{1}{c|}{0.85}     & \multicolumn{1}{c|}{0.65}    & \multicolumn{1}{c|}{\textbf{0.51}} & -48.48\%        \\ \hline
\multicolumn{1}{|c|}{Average}        & \multicolumn{1}{c|}{-}      & \multicolumn{1}{c|}{0.99}      & \multicolumn{1}{c|}{0.73}     & \multicolumn{1}{c|}{0.79}       & \multicolumn{1}{c|}{0.64}         & \multicolumn{1}{c|}{\textbf{0.51}}                   &    -48.98\%               \\ \hline

\end{tabular}}
\label{table_obj}
\end{table*}

\section{Experiments}
\subsection{Datasets}\label{sec::dataset} 
\noindent\textbf{Dataset.}\label{sec:dataset} We validate the effective of our algorithm using datasets from two different scenarios: 

\noindent \textbf{SemanticKITTI \cite{behley2019semantickitti}} is a LiDAR dataset for autonomous driving that contains 22 sequences and 43,552 scans captured by a Velodyne HDL-64E LiDAR sensor. We utilized sequences 00-10 for training and sequences 11-21 for testing lossy compression.

\noindent \textbf{MPEG 8i \cite{MPEG8i}} and \textbf{MVUB \cite{MVUB}} are two object point cloud datasets. MPEG 8i consists of human-shaped point clouds with 10-bit and 12-bit precision, while MVUB contains upper body point cloud sequences with 9-bit and 10-bit precision. We utilized the Soldier10 and Longdress10 point clouds from MPEG 8i, and Andrew10, David10, and Sarah10 from MVUB for training, while the remaining point clouds were used for testing lossless compression.  

\subsection{Experimental Details}\label{sec::detail} 
\noindent\textbf{Experimental Setup.}\label{sec:setup}
We implemented our model using PyTorch and conducted all experiments on a machine equipped with dual NVIDIA GeForce RTX 3090 GPUs. During training, we employed a batch size of 128. The learning rate was set to 1e-4, and the AdamW optimizer was used with its default settings. The neighborhood point cloud size $K$ for the parent node was set to 1024.

\subsection{Experiment Results}\label{sec::result}  
\noindent\textbf{Evaluation Metrics.}\label{sec:metrics} 
We use the following metrics to evaluate both compression performance and the quality of the reconstructed point cloud: 1) Bits per point (Bpp), which assesses compression performance and is applied across all datasets. 2) Point-to-point PSNR (D1 PSNR) \cite{mekuria2017performance} and point-to-plane PSNR (D2 PSNR) \cite{tian2017geometric} to measure the quality of the decoded point clouds. As we focus on the lossless compression task for object point clouds, these metrics are evaluated solely on LiDAR point clouds. 3) Chamfer Distance (CD) \cite{huang20193d, fan2017point}, which is also evaluated on LiDAR point clouds. For the first two metrics, higher values indicate better performance, whereas for the Chamfer Distance, lower values are preferred.

\noindent\textbf{Results.}\label{sec:results} We first evaluate the lossy compression performance of our method against OctAttention \cite{fu2022octattention} and VoxelContext-Net \cite{que2021voxelcontext} on the SemanticKITTI dataset \cite{behley2019semantickitti}. As shown in Fig.~\ref{Fig:FIGKITTI}, our method demonstrates superior rate-distortion performance across all bitrates. Specifically, compared to G-PCC \cite{gpcc}, our approach achieves a 37.95\% reduction in bitrate on SemanticKITTI. At higher bitrates, our method further reduces the bitrate by 9.8\% compared to OctAttention \cite{fu2022octattention}. Unlike VoxelContext-Net \cite{que2021voxelcontext}, which relies solely on voxel data from parent nodes, our method leverages both point cloud and voxel joint contexts. This integration addresses the sparsity inherent in LiDAR point clouds, thereby enhancing spatial perception and improving compression efficiency. 
 
Then, we evaluate the lossless compression performance of our method against G-PCC \cite{gpcc} and OctAttention \cite{fu2022octattention} on the MPEG 8i \cite{MPEG8i} and MVUB \cite{MVUB} datasets. As shown in Tab.~\ref{table_obj}, compared to G-PCC \cite{gpcc}, our method achieves a 36.36\% and 48.98\% reduction in bits per point (bpp) on the MVUB and MPEG 8i datasets, respectively. Additionally, compared to OctAttention \cite{fu2022octattention}, our approach improves compression performance by 7.89\% and 21.3\% on the same datasets. These results demonstrate the effectiveness of our joint context approach, which leverages dense point clouds for precise object surface reconstruction and enhanced compression efficiency. 

\subsection{Ablation Study}   
\noindent\textbf{Effectiveness of Point and Voxel Context.} To investigate the importance of both contexts, we provide an ablation study to verify their effect on the MPEG 8i and MVUB datasets. From Tab.~\ref{point_voxel_alb}, it can be observed that the local voxel information from precursor encoded nodes yielded greater gains compared to using ancestor point cloud information alone. Furthermore, combining both contexts further improved compression performance, underscoring the efficacy of our context fusion approach.

\begin{table}[!h]
\centering
\caption{Ablation study on point and voxel context.}
\resizebox{0.45\textwidth}{!}{%
\setlength{\tabcolsep}{3mm}
\begin{tabular}{|c|c|cc|cc|}
\hline
\multirow{2}{*}{Voxel} & \multirow{2}{*}{Point} & \multicolumn{2}{c|}{Bpp on MVUB}                     & \multicolumn{2}{c|}{Bpp on MPEG 8i}                  \\ \cline{3-6} 
                         &                        & \multicolumn{1}{c|}{Phil10}         & Ricardo10      & \multicolumn{1}{c|}{Loot10}         & Redandblack10  \\ \hline
$\checkmark$             &                        & \multicolumn{1}{c|}{0.911}          & 0.822          & \multicolumn{1}{c|}{0.696}          & 0.838          \\ \hline
                         & $\checkmark$           & \multicolumn{1}{c|}{1.379}          & 1.303          & \multicolumn{1}{c|}{0.990}          & 1.200           \\ \hline
$\checkmark$             & $\checkmark$           & \multicolumn{1}{c|}{\textbf{0.735}} & \textbf{0.652} & \multicolumn{1}{c|}{\textbf{0.482}} & \textbf{0.593} \\ \hline
\end{tabular}%
}
\label{point_voxel_alb}
\end{table}

\begin{figure}[!h]
\centering 
 \subfigure {\includegraphics[scale=0.26]{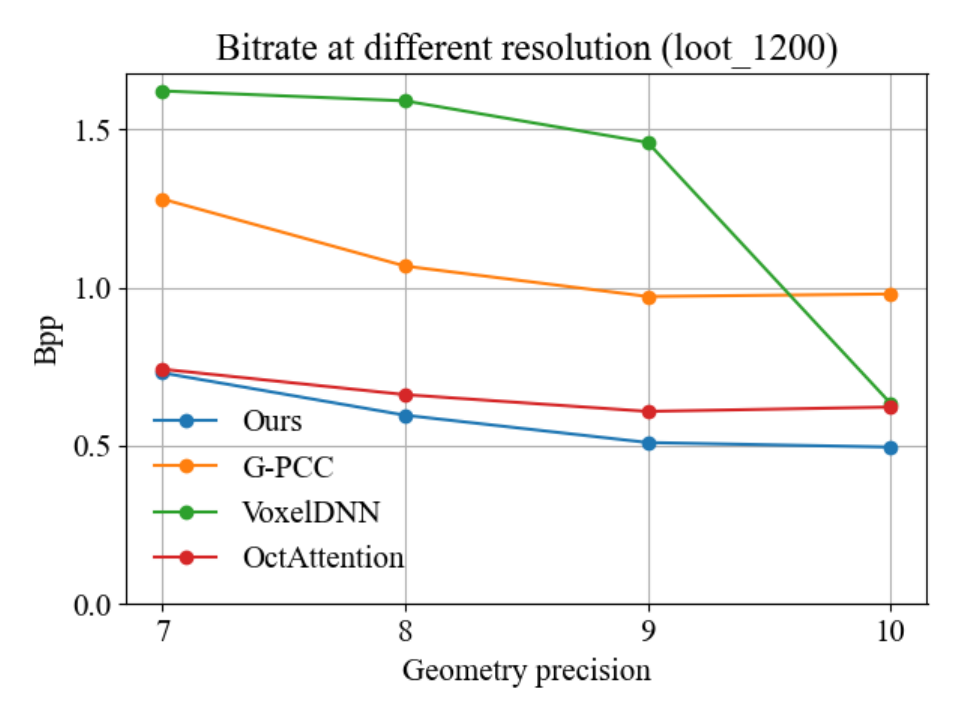}}
 \subfigure {\includegraphics[scale=0.26]{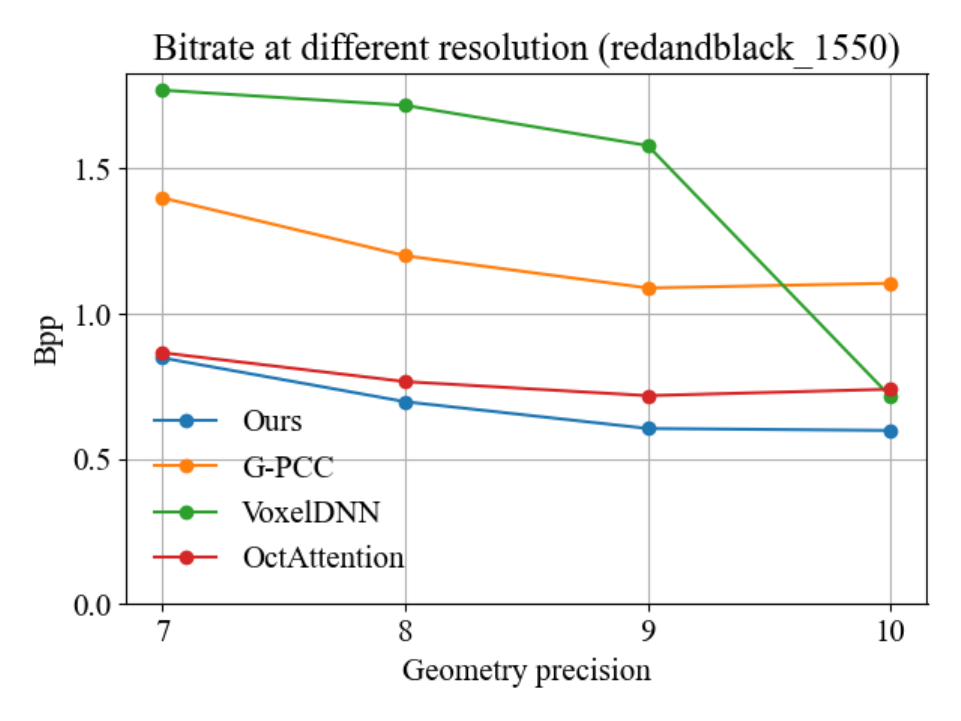}} 
 \caption{Performance at different geometry precision.}
\label{Fig:multiprecision}
\end{figure}

\noindent\textbf{Performance at different geometry precision.} To assess the effectiveness of our method across different resolutions, we compared its compression performance against other methods at varying precisions. As shown in Fig.~\ref{Fig:multiprecision}, our approach consistently outperforms others, with the performance gap widening as point cloud accuracy increases. These results highlight the robustness of our method to different point cloud scales and demonstrate the efficacy of our hierarchical strategy.

\section{Conclusions}
In this work, we propose a novel context called PVContext for point cloud geometric compression. Our hybrid context consists of data from two different modalities: point cloud and voxel. The parent nodes of the target node being encoded form the point cloud context, providing global spatial information. The precursor sibling nodes of the target node form the voxel context, offering more detailed local structural information.
Building on this foundation, we propose a novel entropy model that effectively integrates the aforementioned context features from different modalities and scales. Experimental results demonstrate that our method achieves superior compression performance for both sparse LiDAR point clouds and dense object point clouds.

\bibliographystyle{IEEEtran}
\bibliography{IEEEabrv,refs}

\end{document}